\documentclass[10pt, a4paper]{article}
\PassOptionsToPackage{sort&compress}{natbib}
\usepackage{microtype}
\usepackage{amsmath}
\usepackage{booktabs}
\usepackage{tabularx}
\usepackage{threeparttable}
\usepackage{placeins} 
\usepackage{multirow}
\usepackage{kotex}
\usepackage{makecell}
\usepackage[normalem]{ulem}  
\usepackage{caption}
\usepackage[table]{xcolor}
\usepackage{verbatim}
\usepackage{pgfplots}
\pgfplotsset{compat=1.18}
\usepackage{xurl} 
\usepackage{siunitx}
\usepackage{stfloats}
\usepackage[most]{tcolorbox}
\usepackage[final]{lrec2026}
\usepackage{listings}

\newtcblisting{promptbox}{
  verbatim,
  enhanced,
  breakable,
  listing only,
  colback=white,
  colframe=black!40,
  boxrule=0.6pt,
  arc=1pt,
  left=8pt,right=8pt,top=6pt,bottom=6pt,
  width=\linewidth,
  listing options={
    basicstyle=\ttfamily\small,
    columns=fullflexible,
    keepspaces=true,
    breaklines=true,
    breakautoindent=false, 
    breakindent=0pt,       
    showstringspaces=false,
    frame=none,
    numbers=none
  }
}

\newcolumntype{Y}{>{\raggedright\arraybackslash}X}
\newcolumntype{C}{>{\centering\arraybackslash}p{1.05cm}}

\title{Steering LLMs toward Korean Local Speech:\\%
\texorpdfstring{Iterative Refinement Framework for Faithful Dialect Translation}
{Iterative Refinement Framework for Faithful Dialect Translation}}

\name{Keunhyeung Park, Seunguk Yu, Youngbin Kim} 

\address{Chung-Ang University, Seoul, Republic of Korea\\
         synoark99@cau.ac.kr, seungukyu@gmail.com, ybkim85@cau.ac.kr\\}

\abstract{
Standard-to-dialect machine translation remains challenging due to a persistent dialect gap in large language models and evaluation distortions inherent in n-gram metrics, which favor source copying over authentic dialect translation. In this paper, we propose the dialect refinement (\textbf{\texttt{DIA-REFINE}}) framework, which guides LLMs toward faithful target dialect outputs through an iterative loop of translation, verification, and feedback using external dialect classifiers. To address the limitations of n-gram-based metrics, we introduce the dialect fidelity score (\textbf{DFS}) to quantify linguistic shift and the target dialect ratio (\textbf{TDR}) to measure the success of dialect translation. Experiments on Korean dialects across zero-shot and in-context learning baselines demonstrate that \texttt{DIA-REFINE} consistently enhances dialect fidelity. The proposed metrics distinguish between \textbf{False Success} cases, where high n-gram scores obscure failures in dialectal translation, and \textbf{True Attempt} cases, where genuine attempts at dialectal translation yield low n-gram scores. We also observed that models exhibit varying degrees of responsiveness to the framework, and that integrating in-context examples further improves the translation of dialectal expressions. Our work establishes a robust framework for goal-directed, inclusive dialect translation, providing both rigorous evaluation and critical insights into model performance.
\\ \newline  }

\begin{document}

\maketitleabstract
\section{Introduction}\label{sec:1}

Recent advancements in large language models (LLMs) have revolutionized the field of machine translation~\citep{lyu-etal-2024-paradigm}, highlighting the importance of inclusive approaches for diverse low-resource languages~\citep{CostaJussa2024Scaling200}. In this context, dialect research holds significant academic and social value, as it contributes to preserving the linguistic heritage of potentially marginalized regional languages and enhancing technological inclusivity~\citep{ziems-etal-2022-value}.

Despite the progress of LLMs, a significant performance disparity persists between standard and non-standard dialects~\citep{faisal-etal-2024-dialectbench,fleisig-etal-2024-linguistic,kantharuban-etal-2023-quantifying}. This behavior exhibits an asymmetry depending on the translation direction, with translating from a standard language to a dialect being more challenging than the reverse~\citep{zheng-etal-2022-improving,park-etal-2020-jejueo}. The difficulty of LLMs in translating distinctive dialectal features is evidenced in our experimental results, where some models achieved a zero-shot success rate of only 2\%, with most outputs closely resembling the standard language. While dialect machine translation exhibits these limitations, research on controlling and evaluating the dialectal fidelity of translation outputs remains scarce.

\begin{figure}[t!]
    \centering
    \includegraphics[width=\columnwidth]{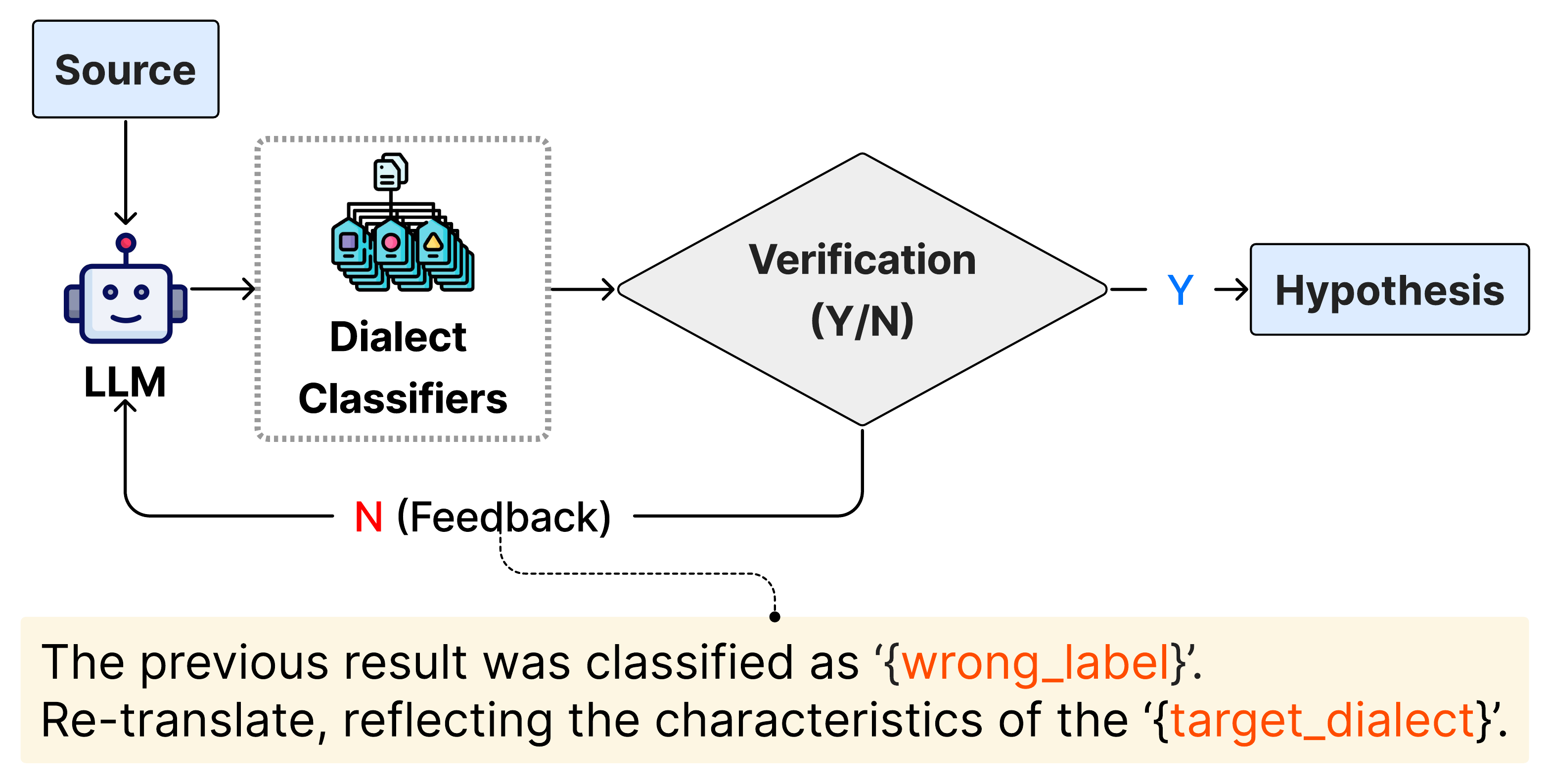} 
    \caption{An overview of our dialect refinement (\texttt{DIA-REFINE}) framework. The LLM's output is verified by an external ensemble of dialect classifiers, which provides explicit feedback to guide the model toward translating the target dialect.}
    \label{fig:framework}
\end{figure}

To address this gap, we propose a framework that utilizes external feedback from dialect classifiers to guide LLMs toward translating target dialect outputs, as illustrated in Figure~\ref{fig:framework}. Our core methodology, dialect refinement (\textbf{\texttt{DIA-REFINE}}), verifies whether the output possesses linguistic features of the target dialect based on the dialect classifier's prediction. If the output fails to satisfy the target dialect condition, explicit feedback derived from the classification result is provided to the model for re-translation. Through this iterative process, the LLM can be effectively controlled to perform goal-oriented dialect translation.

\begin{table*}[t]
\centering
\small
\setlength{\tabcolsep}{4pt}
\renewcommand{\arraystretch}{1.15}

\begin{tabularx}{0.95\textwidth}{@{}p{2.8cm} YY c c c@{}}
\hline
Variants & Forms & Analysis / Key Features & BLEU & chrF++ & \textbf{DFS} \\
\hline

\makecell[tl]{\textbf{Source}\\(Standard)} &
제주도는 특별한 약초들이 많이 없으니까 그냥 그냥 넘긴 거 같아 &
Standard grammar and vocabulary. & -- & -- & -- \\

\makecell[tl]{\textbf{Reference}\\(\textit{Jeju} Dialect)} &
제주도는 특별한 약초들이 많이 엇이난 그냥 그냥 넘긴 거 닮아 &
Contains target dialectal features. & -- & -- & -- \\ \hline

\textbf{Hypothesis 1} &
제주도\underline{엔} 특별한 약초\underline{덜}이 많이 \underline{읎응께} 그냥 그냥 넘긴 거 같아 &
\makecell[tl]{\textbf{True Attempt}: Generates \\ multiple dialectal features \\ 
(\underline{-엔} (-en), \underline{-덜} (-deol), \\ 
\underline{-읎응께} (-eup-eung-kke))} &
27.78 & 36.80 & \textbf{0.672} \\

\textbf{Hypothesis 2} &
제주도는 특별한 약초들이 많이 없으니깐 그냥 그냥 넘긴 거 같아 &
\textbf{False Success}: Output mirrors the standard source rather than the intended dialect. &
\textbf{52.54} & \textbf{67.82} & -0.672 \\
\hline
\end{tabularx}


\small \caption{Evaluation distortions in Korean dialect translation: while n-gram metrics favor surface overlap (\textbf{False Success}), the proposed DFS metric accurately captures the shift toward the dialectal output (\textbf{True Attempt}). Underlined morphemes denote dialectal expressions in the hypothesis. The shared meaning is `\textit{I think they just passed on Jeju Island since it doesn't have a lot of special medicinal herbs there.}'}
\label{tab:false_success_analysis}
\end{table*}

However, traditional n-gram-based metrics such as BLEU~\citep{papineni-etal-2002-bleu} and chrF++~\citep{popovic-2017-chrf} have been repeatedly reported to have a structural blind spot~\citep{aepli-etal-2023-benchmark,chen-etal-2024-iterative}. This vulnerability can lead to higher scores for \textbf{False Success}, which involves merely copying the source, than for a \textbf{True Attempt} to generate dialectal features\footnote{In this paper, we define \textbf{False Success} as cases exhibiting relatively high n-gram but low DFS/TDR scores, and \textbf{True Attempt} as cases showing the opposite pattern, with low n-gram and high DFS/TDR scores, within our comparative analysis across models.}. As shown in Table~\ref{tab:false_success_analysis}, this paradox causes outputs that merely copy the standard source to be evaluated more favorably than genuine attempts to generate dialectal features. This biased evaluation system hinders the accurate assessment of practical gains of the dialectal translation outputs.

To address the limitations of existing metrics and rigorously validate our framework, we introduce a new multifaceted evaluation system. It comprises the dialect fidelity score (\textbf{DFS}), which quantifies the linguistic shift of translated outputs toward the dialectal reference, and the target dialect ratio (\textbf{TDR}), which measures the proportion of outputs correctly classified as the target dialect. These metrics are designed to reliably assess the effectiveness of our framework in steering translation toward the target dialect. To validate the proposed framework, we conducted experiments translating standard Korean into three major dialects with distinct segmental features, \textit{Jeolla}, \textit{Gyeongsang}, and \textit{Jeju}, two of which have been largely unaddressed in previous research. Using several state-of-the-art LLMs, we performed a comparative analysis of existing approaches, including zero-shot and in-context learning, against our proposed \texttt{DIA-REFINE} framework to evaluate model performance under various conditions. The main contributions of this study are as follows: 
\begin{enumerate}
    \item[\textbf{(i)}] We propose the dialect refinement (\textbf{\texttt{DIA-REFINE}}) framework, an effective mechanism for controlling dialect translation.
    \item[\textbf{(ii)}] We introduce a robust evaluation system with \textbf{DFS} and \textbf{TDR} metrics, addressing the limitations of n-gram-based metrics and enabling an accurate assessment of dialectal fidelity.
    \item[\textbf{(iii)}] Our analysis demonstrates that the proposed metrics effectively distinguish between cases of \textbf{False Success} and \textbf{True Attempt} across models. Moreover, we establish that combining our framework with in-context examples yields the most effective performance in dialect translation.
\end{enumerate}

This paper is organized as follows: Section~\ref{sec:2} reviews related work on recent trends in dialect research, steering generation and challenges in dialect evaluation. Section~\ref{sec:3} and~\ref{sec:4} present the \texttt{DIA-REFINE} framework and experimental setup, respectively. Section~\ref{sec:5} details our evaluation metrics. Section~\ref{sec:6} reports the main results and our detailed analysis, and Section~\ref{sec:7} concludes. We plan to release the code to facilitate the proposed framework and metrics, thereby supporting further research in this area\footnote{Due to the source dataset's terms of use, the processed dataset cannot be released; however, we made our code publicly available to support reproducibility. Available at: \url{https://github.com/keunhyeung/DIA-REFINE}}.

\section{Related Work}\label{sec:2}

\subsection{Recent Trends in Dialect Research}\label{sec:2.1}

NLP research has traditionally focused on a small number of standard, high-resource languages~\citep{ziems-etal-2022-value,nigatu-etal-2024-zenos}. This has created a significant performance disparity between standard and non-standard varieties. For instance, tweets in African American English are up to twice as likely to be mislabeled as offensive~\citep{sap-etal-2019-risk}, and models like ChatGPT produce responses to non-standard dialects that are perceived as more stereotypical and demeaning ~\citep{fleisig-etal-2024-linguistic}. These biases perpetuate linguistic discrimination and marginalize speakers of non-standard dialects.

To address these limitations, a wider variety of dialects has received increasing attention in recent research. These efforts have led to the creation of notable resources, such as the parallel data MADAR for Arabic~\citep{bouamor-etal-2018-madar} and YORÙLECT for Yorùbá~\citep{ahia-etal-2024-voices}. ARGEN benchmark offers a comprehensive framework for evaluating dialect generation~\citep{nagoudi-etal-2022-arat5}. The latest benchmark WMT24++ includes 10 dialects across 5 languages~\citep{deutsch-etal-2025-wmt24}. Collectively, these efforts reflect a trend toward addressing a broader range of language varieties and underscore the need for dialect-specific research. In contrast to this trend, research on Korean has been limited, focusing almost exclusively on the \textit{Jeju} dialect through efforts to construct the JIT dataset~\citep{park-etal-2020-jejueo} and cross-lingual pre-training with Japanese ~\citep{zheng-etal-2022-improving}. Our work is significant in that it expands this scope to include the \textit{Jeolla} and \textit{Gyeongsang} dialects, which have been largely unaddressed in previous Korean dialect research.

\subsection{Generation Control using External Feedback}\label{sec:2.2}


Leveraging external information to control model outputs toward specific goals is an active line of research. Classifier-guided diffusion steers the sampling process through the gradient of a pre-trained classifier, thereby improving the quality of the generated samples~\citep{dhariwal2021diffusion}. SELF-REFINE iterates a generate–feedback–refine loop with a single model and requires no supervised data~\citep{madaan2023selfrefine}. In machine translation tasks, iteratively prompting an LLM to refine its result can improve the fluency and naturalness itself, mimicking a human-like editing process~\citep{chen-etal-2024-iterative}. 


In this paper, we extend the idea of iterative refinement by using external feedback to control LLM output quality for the dialect translation task. In our \texttt{DIA-REFINE} framework, the dialect classifier functions as a linguistic tool to verify the dialectal fidelity of the model's output. We explicitly utilized the feedback from the high-performance ensemble model for a task-specific control that guides the dialect translation process.


\subsection{Distortion in Dialect Translation Evaluation}\label{sec:2.3}

Despite the widespread adoption of embedding-based metrics for evaluating machine translation tasks~\citep{larionov-etal-2024-xcomet}, these advanced metrics show a significant lack of dialect robustness~\citep{sun-etal-2023-dialect}. Therefore, research on standard-to-dialect translation has still heavily relied on n-gram metrics such as BLEU and chrF~\citep{faisal-etal-2024-dialectbench,zheng-etal-2022-improving,nagoudi-etal-2022-arat5,ahia-etal-2024-voices}. 

However, these metrics can distort evaluation by overweighting surface-form overlap rather than translation quality. \citet{aepli-etal-2023-benchmark} showed that for Swiss German, a dialect without standardized orthography, semantically incorrect yet lexically similar outputs often scored higher on n-gram metrics than correct ones. In this aspect, \citet{chen-etal-2024-iterative} report that, in multilingual translation tasks, iterative translation refinement tends to reduce n-gram–based scores even as human preferences increase, revealing a divergence between n-gram–based and semantic evaluations.


This \textbf{False Success} phenomenon has been reported across dialect studies. \citet{park-etal-2020-jejueo} found that a trivial copy-model achieved high BLEU scores on translation from standard Korean to the \textit{Jeju} dialect. Similarly,~\citet{liu-2022-low} found that BLEU can systematically favor copy-bias models in Cantonese translation. Analyzing multi-dialect performance,~\citet{kantharuban-etal-2023-quantifying} quantified a strong correlation between dialect–standard lexical similarity and performance on n-gram metrics. Collectively, these studies show that relying solely on n-gram metrics for standard-to-dialect evaluation can severely distort assessment. To address this evaluation gap, we propose two complementary metrics, DFS and TDR, to enable a more accurate measure of dialectal fidelity.

\begin{table}[t]
\centering
\small
\begin{tabular}{lccc}
\toprule
Class & Precision & Recall & F1 \\
\midrule
\textit{Jeolla}        & 0.9308 & 0.9150 & 0.9228 \\
\textit{Gyeongsang}    & 0.9135 & 0.9080 & 0.9107 \\
\textit{Jeju}          & 0.9888 & 0.9720 & 0.9803 \\
\textit{Standard}      & 0.9188 & 0.9510 & 0.9346 \\
\textit{Unknown}       & 0.9950 & 1.0000 & 0.9975 \\
\bottomrule
\end{tabular}

\vspace{1.5ex} 

\begin{tabular}{lc}
\toprule
Overall Metric & Score \\
\midrule
Accuracy       & 0.9492 \\
Macro F1-Score & 0.9494 \\
\bottomrule
\end{tabular}

\small \caption{Performance of the best ensemble dialect classifier from our experiments, with the top showing per-class results and the bottom table showing the overall average.}
\label{tab:ensemble_cls}
\end{table}

\begin{table*}[t]
\renewcommand{\arraystretch}{1.5}
\centering
\resizebox{0.9\textwidth}{!}{%
\begin{tabular}{lcccc}
\hline
Method            & \multicolumn{1}{l}{Candidates} & Selection              & Verification         & Feedback Loop     \\ \hline
Baseline (ZS, ICL) & $1$                              & N/A                    & N/A                  & N/A               \\
Baseline + \texttt{DIA-REFINE(S)} & $1$                              & N/A                    & $\mathcal{C}(y) = d_{\mathrm{tgt}}$  & Retry on mismatch \\
Baseline + \texttt{DIA-REFINE(M)} & $k \in \{3, 4, 5\}$                & Select top-1 as $y^*$ & $\mathcal{C}(y^*) = d_{\mathrm{tgt}}$ & Retry on mismatch \\
\hline
\end{tabular}
}
\small \caption{Variants of the proposed \texttt{DIA-REFINE} differing in candidate generation and hypothesis selection strategies. The framework extends the baseline with verification and feedback loops, allowing up to two retries. For \texttt{DIA-REFINE(M)}, the best candidate $y^*$ is selected from $k$ options by maximizing their posterior probability. Throughout this process, we employed our trained ensemble dialect classifier $\mathcal{C}$.}
\label{tab:cil_framework}
\end{table*}

\section{Proposed Methodology: The DIA-REFINE Framework}\label{sec:3}
We propose dialect-refinement (\textbf{\texttt{DIA-REFINE}}), a framework designed to steer LLMs toward consistent and high-fidelity dialect translation. At its core, \texttt{DIA-REFINE} operates through an iterative loop of translation and verification, using feedback from external  dialect classifiers to progressively guide the translation output toward the target dialect.

\subsection{Dialect Classifier Dataset Construction}\label{sec:3.1}
We collected the Korean dialect data corpus~\citeplanguageresource{aihub:dialectdata}, a public resource to preserve endangered dialects. It is a parallel corpus of standard Korean and dialectal sentence pairs from five major regions: \textit{Gangwon}, \textit{Chungcheong}, \textit{Jeolla}, \textit{Gyeongsang}, and \textit{Jeju}.

From this corpus, we constructed a dialect classifier dataset. We selected \textit{Jeolla}, \textit{Gyeongsang}, and \textit{Jeju} as the target dialects due to their salient features\footnote{The \textit{Jeolla} and \textit{Gyeongsang} dialects are clearly distinct from standard Korean, while the \textit{Jeju} dialect has a unique feature owing to its geographically isolated location. In contrast, the \textit{Gangwon} and \textit{Chungcheong} dialects are generally considered variants of standard Korean, exhibiting fewer linguistic differences.}. These were supplemented by \textit{Standard} and \textit{Unknown} classes, representing no dialect and mixed-dialect sentences. We compiled 10,000 samples per class to prioritize robustness, resulting in 50,000 total samples. We divided each class into a 9:1 ratio for training and evaluation samples. Class-specific curation is as follows:
\begin{itemize}
    \item \textbf{The \textit{Dialect} classes} (\textit{Jeolla}, \textit{Gyeongsang}, and \textit{Jeju}) include only samples where the normalized Levenshtein distance~\citep{10.1109/TPAMI.2007.1078} between the standard source and its dialect counterpart is $\ge$ 0.1, ensuring form-level divergence~\citep{johannessen-etal-2020-comparing}.
    \item \textbf{The \textit{Standard} class} comprises no dialect samples drawn at random from the source.
    \item \textbf{The \textit{Unknown} class} is constructed as a hard-negative set to make the dialect classifier more robust~\citep{li-etal-2024-generating}. We first extracted salient dialect lexical features using TF-IDF and then prompted Gemini-2.5-Flash-Lite\footnote{\label{gemini}\url{https://aistudio.google.com/}} to generate sentences mixing distinct dialects that do not conform to a single dialect. This process was conducted to prevent overfitting and avoid ambiguous outputs that LLMs may produce.

\end{itemize}

\subsection{Building an Ensemble of Dialect Classifiers}\label{sec:3.2}
We adopted an ensemble approach~\citep{arango2024ensembling} to enhance robustness and stability of the dialect classifier $\mathcal{C}$. We fine-tuned 
five distinct Korean pre-trained language models\footnote{\label{plm1}\url{https://huggingface.co/klue/bert-base}}\footnote{\label{plm2}\url{https://huggingface.co/klue/roberta-large}}\footnote{\label{plm3}\url{https://huggingface.co/BM-K/KoSimCSE-roberta}}\footnote{\label{plm4}\url{https://huggingface.co/monologg/koelectra-base-v3-discriminator}}\footnote{\label{plm5}\url{https://huggingface.co/beomi/KcELECTRA-base}} using a set of shared hyperparameters, with a batch size of 16, a learning rate of 4e-5, a weight decay of 0.01, 4 training epochs, a maximum input length of 64, and a seed of 1337 for reproducibility.

After fine-tuning each model, we searched for the optimal ensemble scenario among the 31 combinations. While the lowest-performing case yielded an accuracy of 92.92\%, we found that \texttt{BM-K/KoSimCSE-roberta}\footref{plm3} combined with \texttt{beomi/KcELECTRA-base}\footref{plm5} performed best. As shown in Table~\ref{tab:ensemble_cls}, this ensemble achieved an overall accuracy of 94.92\% with balanced performance.

\subsection{DIA-REFINE Framework for Dialect Translation}\label{sec:3.3}
We operate our \textbf{\texttt{DIA-REFINE}} on the trained ensemble dialect classifier. First, an LLM generates a candidate translation $y$ following the given instruction, then $\mathcal{C}$ determines whether the $y$ aligns with the intended target dialect $d_{\textit{tgt}}$. If it was classified as a non-targeted dialect (i.e., $\mathcal{C}(y) \neq d_{\textit{tgt}}$), the framework requests retranslation using a prompt that contains explicit feedback. This feedback goes beyond a simple mismatch signal, specifying which class the output was misclassified into to guide the correction~\citep{dhariwal2021diffusion}.

Furthermore, when the \texttt{DIA-REFINE} detects the model oscillating between different error types on consecutive retries, the prompt explicitly flags this pattern to help the model break the cycle and steer its generation more effectively. As detailed in Table~\ref{tab:cil_framework}, we defined two variants of our framework that primarily differ in their candidate generation and hypothesis selection strategies:
\begin{itemize}
    \item \textbf{\texttt{DIA-REFINE(S)} (Single-candidate)} follows a simple single-candidate approach that verifies each output as it is generated.
    \item \textbf{\texttt{DIA-REFINE(M)} (Multi-candidates)} generates $k$ outputs and selects the most promising candidate $y^*$ based on the dialect classifier's posterior probability. Only this selected candidate is then verified. This generate-and-select approach aims to broaden the search space, with $k$ starting at 3 and incrementing by one with each retry to enhance exploration.
\end{itemize}
Both variants allow up to two retries on mismatch, for a total of three attempts.

\section{Experimental Setup}\label{sec:4}

\subsection{Dialect Translation Test Set}\label{sec:4.1}

From the corpus in Section~\ref{sec:3.1}, we created a dialect translation test set to evaluate translation performance. For this set, we randomly sampled 300 pairs each from the \textit{Jeolla}, \textit{Gyeongsang}, and \textit{Jeju} dialects, which exhibit salient features. We only included sentences with a length of $\geq$ 30 characters to ensure they contained sufficient content.

\subsection{Models and Baselines}\label{sec:4.2}

Given our focus on Korean dialects, our evaluation includes the three Korean-specialized models HyperCLOVAX\footnote{\url{https://huggingface.co/naver-hyperclovax/HyperCLOVAX-SEED-Text-Instruct-1.5B}}, EXAONE-3.5\footnote{\url{https://huggingface.co/LGAI-EXAONE/EXAONE-3.5-7.8B-Instruct}}, and EEVE\footnote{\url{https://huggingface.co/yanolja/YanoljaNEXT-EEVE-Instruct-10.8B}}. To provide a broader comparative context, we also evaluated an open general-purpose model Llama-3.1\footnote{\url{https://huggingface.co/meta-llama/Llama-3.1-8B-Instruct}}, and a proprietary model Gemini-1.5\footref{gemini}.

We included baselines with zero-shot (ZS) and in-context learning (ICL). The latter is known to be effective in low-resource translation \citep{pei-etal-2025-understanding}. From the corpus in Section~\ref{sec:3.1}, we prepared an ICL example pool of 10,000 pairs for each class, ensuring no overlap with the test set. We retrieved the top-10 most relevant pairs from the example pool using BM25 to construct ICL examples.
\begin{table}[t]
\centering
\renewcommand{\arraystretch}{1.2}
\resizebox{0.95\linewidth}{!}{%
\begin{tabular}{
  @{} l
  @{\hspace{10pt}} S[table-format=-1.2] @{\hspace{12pt}} S[table-format=-1.2]
  @{\hspace{20pt}} S[table-format=-1.2] @{\hspace{12pt}} S[table-format=-1.2]
  @{}
}
\toprule
\multirow{2}{*}{Model}
& \multicolumn{2}{c}{ZS}
& \multicolumn{2}{c}{ICL+\texttt{DIA-REFINE(M)}} \\
\cmidrule(lr){2-3}\cmidrule(lr){4-5}
& {Vanilla} & {Fine-tuned}
& {Vanilla} & {Fine-tuned} \\
\midrule
HyperCLOVAX & -0.01 & -0.37 & 0.00 &  -0.06 \\
Llama-3.1   & -0.01 & -0.37 & 0.00 &  -0.10\\
EEVE        & -0.01 & -0.38 & 0.00 &  -0.05 \\
EXAONE-3.5  & -0.01 & -0.39 & 0.00 &  0.27 \\
Gemini-1.5  &  0.00 &  0.36 & 0.00 &  0.39 \\
\bottomrule
\end{tabular}%
} 
\small \caption{Averaged DFS scores between vanilla model\footref{plm5} and fine-tuned variants, evaluated on the dialect translation test set. The dialect-aware embeddings yield pronounced scores with sensitivity to dialectal features, whereas the standard embeddings indicate reduced effectiveness.}


\label{tab:table4}
\end{table}

\begin{table*}[t!]
\renewcommand{\arraystretch}{1.55}
\centering
\resizebox{0.85\textwidth}{!}{%
\begin{tabular}{@{}llcccccc@{}}
\hline
Metric & Method & \textbf{HyperCLOVAX} & \textbf{Llama-3.1} & \textbf{EEVE} & \textbf{EXAONE-3.5} & \textbf{Gemini-1.5} & Model Avg \\
\hline

\multirow{6}{*}{\textbf{chrF++}} 
 & ZS         & 60.33 & 68.97 & 54.03 & 29.05 & 39.50 & 50.38 \\

 & \cellcolor{gray!10}ZS + DIA-REFINE(S) & \cellcolor{gray!10}60.76 & \cellcolor{gray!10}62.52 & \cellcolor{gray!10}52.41 & \cellcolor{gray!10}29.94 & \cellcolor{gray!10}38.03 & \cellcolor{gray!10}48.73 \\

 & \cellcolor{gray!15}ZS + DIA-REFINE(M) & \cellcolor{gray!15}52.52 & \cellcolor{gray!15}44.29 & \cellcolor{gray!15}39.82 & \cellcolor{gray!15}25.20 & \cellcolor{gray!15}38.55 & \cellcolor{gray!15}40.08 \\

 & ICL        & 59.17 & 71.31 & 70.99 & 65.41 & 63.08 & 65.99 \\

 & \cellcolor{gray!10}ICL + DIA-REFINE(S) & \cellcolor{gray!10}55.35 & \cellcolor{gray!10}70.42 & \cellcolor{gray!10}70.48 & \cellcolor{gray!10}60.19 & \cellcolor{gray!10}62.13 & \cellcolor{gray!10}63.72 \\

 & \cellcolor{gray!15}ICL + DIA-REFINE(M) & \cellcolor{gray!15}46.00 & \cellcolor{gray!15}66.15 & \cellcolor{gray!15}65.67 & \cellcolor{gray!15}54.95 & \cellcolor{gray!15}61.68 & \cellcolor{gray!15}58.89 \\
\midrule

\multirow{6}{*}{\textbf{BLEU}} 
 & ZS         & 47.64 & 58.53 & 41.71 & 11.44 & 23.06 & 36.48 \\

 & \cellcolor{gray!10}ZS + DIA-REFINE(S) & \cellcolor{gray!10}47.94 & \cellcolor{gray!10}49.82 & \cellcolor{gray!10}39.49 & \cellcolor{gray!10}11.97 & \cellcolor{gray!10}21.92 & \cellcolor{gray!10}34.23 \\

 & \cellcolor{gray!15}ZS + DIA-REFINE(M) & \cellcolor{gray!15}38.82 & \cellcolor{gray!15}28.98 & \cellcolor{gray!15}25.24 & \cellcolor{gray!15}8.13 & \cellcolor{gray!15}22.13 & \cellcolor{gray!15}24.66 \\

 & ICL        & 46.21 & 61.75 & 61.34 & 52.88 & 49.10 & 54.25 \\

 & \cellcolor{gray!10}ICL + DIA-REFINE(S) & \cellcolor{gray!10}41.44 & \cellcolor{gray!10}60.57 & \cellcolor{gray!10}60.68 & \cellcolor{gray!10}45.75 & \cellcolor{gray!10}47.73 & \cellcolor{gray!10}51.23 \\

 & \cellcolor{gray!15}ICL + DIA-REFINE(M) & \cellcolor{gray!15}31.70 & \cellcolor{gray!15}54.67 & \cellcolor{gray!15}54.30 & \cellcolor{gray!15}39.14 & \cellcolor{gray!15}47.22 & \cellcolor{gray!15}45.41 \\
\midrule

\multirow{6}{*}{\textbf{DFS}} 
 & ZS         & -0.37 & -0.37 & -0.38 & -0.39 &  0.36 & -0.23 \\

 & \cellcolor{gray!10}ZS + DIA-REFINE(S) & \cellcolor{gray!10}-0.37 & \cellcolor{gray!10}-0.37 & \cellcolor{gray!10}-0.37 & \cellcolor{gray!10}-0.32 & \cellcolor{gray!10}0.38 & \cellcolor{gray!10}-0.21 \\

 & \cellcolor{gray!15}ZS + DIA-REFINE(M) & \cellcolor{gray!15}-0.32 & \cellcolor{gray!15}-0.32 & \cellcolor{gray!15}-0.28 & \cellcolor{gray!15}-0.08 & \cellcolor{gray!15}0.40 & \cellcolor{gray!15}-0.12 \\

 & ICL        & -0.30 & -0.26 & -0.26 & -0.04 &  0.31 & -0.11 \\

 & \cellcolor{gray!10}ICL + DIA-REFINE(S) & \cellcolor{gray!10}-0.24 & \cellcolor{gray!10}-0.24 & \cellcolor{gray!10}-0.23 & \cellcolor{gray!10}0.07 & \cellcolor{gray!10}0.36 & \cellcolor{gray!10}-0.06 \\

 & \cellcolor{gray!15}ICL + DIA-REFINE(M) & \cellcolor{gray!15}-0.06 & \cellcolor{gray!15}-0.10 & \cellcolor{gray!15}-0.05 & \cellcolor{gray!15}0.27 & \cellcolor{gray!15}0.39 & \cellcolor{gray!15}0.09 \\
\midrule

\multirow{6}{*}{\textbf{TDR}} 
 & ZS         & 0.03 & 0.03 & 0.02 & 0.01 & 0.93 & 0.20 \\

 & \cellcolor{gray!10}ZS + DIA-REFINE(S) & \cellcolor{gray!10}0.04 & \cellcolor{gray!10}0.03 & \cellcolor{gray!10}0.03 & \cellcolor{gray!10}0.08 & \cellcolor{gray!10}0.98 & \cellcolor{gray!10}0.23 \\

 & \cellcolor{gray!15}ZS + DIA-REFINE(M) & \cellcolor{gray!15}0.08 & \cellcolor{gray!15}0.08 & \cellcolor{gray!15}0.13 & \cellcolor{gray!15}0.42 & \cellcolor{gray!15}0.99 & \cellcolor{gray!15}0.34 \\

 & ICL        & 0.09 & 0.15 & 0.16 & 0.41 & 0.85 & 0.33 \\

 & \cellcolor{gray!10}ICL + DIA-REFINE(S) & \cellcolor{gray!10}0.15 & \cellcolor{gray!10}0.18 & \cellcolor{gray!10}0.19 & \cellcolor{gray!10}0.60 & \cellcolor{gray!10}0.97 & \cellcolor{gray!10}0.42 \\

 & \cellcolor{gray!15}ICL + DIA-REFINE(M) & \cellcolor{gray!15}0.41 & \cellcolor{gray!15}0.36 & \cellcolor{gray!15}0.42 & \cellcolor{gray!15}0.94 & \cellcolor{gray!15}0.99 & \cellcolor{gray!15}0.62 \\



\hline
\end{tabular}
}
\small \caption{Performance comparison of five language models across six methodologies, evaluated with four key metrics. All scores are averaged over three target dialects (\textit{Jeolla}, \textit{Gyeongsang} and \textit{Jeju}). The Model Avg column presents the average performance across all models for each method.}
\label{tab:full_results}
\end{table*}

\section{Evaluation Framework}\label{sec:5}

\subsection{Limitations of N-gram Metrics}\label{sec:5.1}
By overweighting surface-form similarity rather than dialect fidelity, n-gram metrics can award high scores to outputs to the standard form, yielding \textbf{False Success}~\citep{park-etal-2020-jejueo,liu-2022-low}. For instance, Hypothesis 2 in Table~\ref{tab:false_success_analysis}, which largely mirrors the standard source, achieves deceptively high scores of 52.54 BLEU and 67.82 chrF++. In contrast, Hypothesis 1 makes \textbf{True Attempt} by introducing dialect features like \texttt{-엔 (-en)}, \texttt{-덜 (-deol)}, and \texttt{-읎응께 (-eup-eung-kke)}. Although these features do not belong to the target \textit{Jeju} dialect, this genuine effort is penalized with significantly lower scores of 27.78 BLEU and 36.80 chrF++. This example demonstrates that relying solely on n-gram metrics is not only insufficient but also potentially misleading, thereby making them unsuitable for properly validating the effectiveness of the \texttt{DIA-REFINE} framework.

\subsection{Proposed Metrics: DFS and TDR}\label{sec:5.2}
To address this, we complement n-gram metrics with our two proposed metrics. First, \textbf{dialect fidelity score (DFS)} is a continuous measure of whether the hypothesis $h$ is linguistically closer to the dialectal reference $r$ than to the standard source $s$. In an embedding space, we compute the log ratio of their cosine similarities, as shown in Eq.~\eqref{eq:dfs}. We add $1$ to the cosine similarity and a small constant $\varepsilon$ ($=10^{-6}$) to ensure numerical stability for the logarithm. A positive DFS indicates a shift toward the dialect, while a negative value indicates proximity to the standard variety.
\begin{equation}
\label{eq:dfs}
\mathrm{DFS}
= \log\!\left(
\frac{\,1 + \cos\!\big(\mathbf{e}_{h}, \mathbf{e}_{r}\big) + \varepsilon\,}
     {\,1 + \cos\!\big(\mathbf{e}_{h}, \mathbf{e}_{s}\big) + \varepsilon\,}
\right),
\end{equation}

To ensure reliable representations, we extracted dialect-aware embeddings $\mathbf{e}_h, \mathbf{e}_r, \mathbf{e}_s$ using our fine-tuned \texttt{beomi/KcELECTRA-base} model, with the final classification layer removed. As shown in Table~\ref{tab:table4}, embeddings from the dialect-aware model yielded reasonable DFS scores compared to the vanilla model. While our model exhibits a score range from -0.39 to 0.27 across methods, the vanilla model shows lower sensitivity, varying only from -0.01 to 0. This demonstrates that our dialect-aware model yields an embedding space that truly reflects the intended dialectal features.


The effectiveness of this metric is apparent when returning to the examples in Table~\ref{tab:false_success_analysis}. The \textbf{True Attempt} of Hypothesis 1, despite its low n-gram scores, yields a high positive DFS of 0.672, correctly capturing its linguistic deviation from the standard source. Conversely, the \textbf{False Success} of Hypothesis 2 yields a negative DFS of -0.672, indicating its close alignment with the standard source. In our low-resource language setting using dialects, where embedding-based metrics are previously constrained~\citep{mukherjee-etal-2025-high}, DFS highlights the potential of an embedding-based evaluation. However, it is important to note that DFS is not a standalone measure of overall translation success and should therefore be interpreted in conjunction with other metrics.

Second, \textbf{target dialect ratio (TDR)} represents the proportion of outputs correctly generated in the target dialect. For a set of generated hypotheses $H$, it is the proportion of the hypotheses classified by $\mathcal{C}$ as the target dialect $d_{\textit{tgt}}$, as shown in Eq.~\eqref{eq:tdr}. This metric provides a clear quantification of the actual success rate in dialect translation.
\begin{equation}
\label{eq:tdr}
\mathrm{TDR}
= \frac{|\{y \in H \mid \mathcal{C}(y) = d_{\mathrm{tgt}}\}|}{|H|}.
\end{equation}

We used DFS and TDR to mitigate the limitations of n-gram-based metrics and the absence of reliable dialectal embedding-based metrics.

\pgfplotsset{compat=1.18} 

\begin{figure}[t]
  \small
  \centering
  \begin{tikzpicture}
    \pgfplotsset{
        width=0.85\linewidth, 
        height=7.2cm,
        symbolic x coords={ZS, ZS+DIA-REFINE(S), ZS+DIA-REFINE(M), ICL, ICL+DIA-REFINE(S), ICL+DIA-REFINE(M)},
        xtick=data,
        x tick label style={rotate=45, anchor=east, font=\footnotesize},
        ymin=0, ymax=100,
        grid=major,
        grid style={dotted, gray!50}, 
        label style={font=\footnotesize},
        tick label style={font=\footnotesize},
        every axis plot/.style={line width=1.0pt, mark size=2.2pt} 
    }

    \begin{axis}[
        axis y line*=left,
        ylabel={BLEU $\uparrow$},
        legend style={
            at={(0.02,0.98)}, 
            anchor=north west,
            draw=black, 
            fill=white, 
            font=\scriptsize,
        },
    ]
      \addplot[red, dashed, mark=x]
        coordinates {
          (ZS,47.64) (ZS+DIA-REFINE(S),47.94) (ZS+DIA-REFINE(M),38.82)
          (ICL,46.21) (ICL+DIA-REFINE(S),41.44) (ICL+DIA-REFINE(M),31.70)
        };

      \addplot[blue, dashed, mark=x]
        coordinates {
          (ZS,11.44) (ZS+DIA-REFINE(S),11.97) (ZS+DIA-REFINE(M),8.13)
          (ICL,52.88) (ICL+DIA-REFINE(S),45.75) (ICL+DIA-REFINE(M),39.14)
        };

      \addlegendimage{red, solid, mark=*}
      \addlegendimage{blue, solid, mark=*}
      \addlegendimage{red, solid, mark=*}
      \addlegendimage{blue, dahsed, mark=*}
      
      \legend{HyperCLOVAX (BLEU), EXAONE-3.5 (BLEU), HyperCLOVAX (TDR), EXAONE-3.5 (TDR)}
    \end{axis}

    \begin{axis}[
        axis y line*=right,
        axis x line=none, 
        ylabel={TDR $\uparrow$},
    ]
      \addplot[red, solid, mark=*]
        coordinates {
          (ZS,3) (ZS+DIA-REFINE(S),4) (ZS+DIA-REFINE(M),8)
          (ICL,9) (ICL+DIA-REFINE(S),15) (ICL+DIA-REFINE(M),41)
        };

      \addplot[blue, solid, mark=*]
        coordinates {
          (ZS,1) (ZS+DIA-REFINE(S),8) (ZS+DIA-REFINE(M),42)
          (ICL,41) (ICL+DIA-REFINE(S),60) (ICL+DIA-REFINE(M),94)
        };
    \end{axis}
    
  \end{tikzpicture}
  \small \caption{Divergent trends between BLEU and the proposed TDR metric. This highlights a key limitation of BLEU, as it fails to reflect the positive impact of our methods on dialect translation.}
  \label{fig:bleu-tdr}
\end{figure}

\section{Results and Analysis}\label{sec:6}

\subsection{Unmasking False Success Cases}\label{sec:6.1}

Our results clearly demonstrate the limitations of n-gram metrics and validate the necessity of our proposed metrics. As detailed in Table~\ref{tab:full_results}, in the ZS setting, several models achieved deceptively high n-gram scores despite poor dialect translation. HyperCLOVAX, Llama-3.1, and EEVE achieved high chrF++/BLEU scores of 60.33/47.64, 68.97/58.53, and 54.03/41.71, respectively. However, these scores significantly contrast with their performance on the proposed metrics, with TDRs all $\le$ 0.03 and consistently negative DFS scores around -0.37. This indicates that they defaulted to the standard language in nearly 97\% of cases.

As shown in Figure~\ref{fig:bleu-tdr}, the dashed and solid lines represent the BLEU and TDR scores, corresponding to the method used. While the use of ICL and \texttt{DIA-REFINE(M)} led to a steep increase in TDR scores, the BLEU scores did not show an evident correlation. If BLEU had been sensitive to the changes in the dialectal output brought by each method, the dashed line would have mirrored that of the solid line, at least for a highly effective translation method like ICL.


A clear illustration of this evaluation paradox is provided by Gemini-1.5, which achieved comparatively low chrF++/BLEU scores of 39.50/23.06 in the ZS setting, yet exhibited a notably high TDR of 0.93. Relying solely on n-gram metrics would have mischaracterized Gemini-1.5 as the lowest-performing model, when in fact it was the most successful model. This finding underscores the limitations of n-gram-based traditional metrics and suggests that employing the proposed DFS and TDR offers a more accurate and reliable evaluation of dialectal outputs by revealing the \textbf{False Success} phenomenon in our experiments.

\subsection{Performance Trajectory across Methodologies}\label{sec:6.2}

\begin{table*}[t!]
\renewcommand{\arraystretch}{1.55}
\centering
\resizebox{0.6\textwidth}{!}{%
\small
\setlength{\tabcolsep}{8pt}
\begin{tabular}{@{}l c c c c c@{}}
\toprule
Method & chrF++ & BLEU & \textbf{DFS} & \textbf{TDR} & \textbf{Attempts} \\
\midrule
ZS                  & 63.34 & 51.62 & -0.37 & 0.00 & 1.00 \\
\cellcolor{gray!10}ZS + DIA-REFINE(S) & \cellcolor{gray!10}60.49 & \cellcolor{gray!10}47.59 & \cellcolor{gray!10}-0.36 & \cellcolor{gray!10}0.01 & \cellcolor{gray!10}2.99 \\
\cellcolor{gray!15}ZS + DIA-REFINE(M) & \cellcolor{gray!15}46.44 & \cellcolor{gray!15}32.16 & \cellcolor{gray!15}-0.29 & \cellcolor{gray!15}0.09 & \cellcolor{gray!15}2.95 \\
\midrule
ICL                 & 74.35 & 64.74 & -0.34 & 0.00 & 1.00 \\
\cellcolor{gray!10}ICL + DIA-REFINE(S) & \cellcolor{gray!10}70.99 & \cellcolor{gray!10}60.48 & \cellcolor{gray!10}-0.30 & \cellcolor{gray!10}0.06 & \cellcolor{gray!10}2.95 \\
\cellcolor{gray!15}ICL + DIA-REFINE(M) & \cellcolor{gray!15}62.64 & \cellcolor{gray!15}50.59 & \cellcolor{gray!15}\textbf{-0.12} & \cellcolor{gray!15}\textbf{0.28} & \cellcolor{gray!15}\textbf{2.81} \\
\bottomrule
\end{tabular}
}
\small \caption{Performance comparison on a shared subset of 2,305 difficult samples based on the averaged results across five models. The multi-candidate variant, \texttt{DIA-REFINE(M)}, demonstrates the most pronounced corrective effect in converting initial failures into successful dialect translations.}
\label{tab:difficult_samples}
\end{table*}

As detailed in Table~\ref{tab:full_results}, the models exhibited distinct performance trajectories. Gemini-1.5, in particular, demonstrated a pronounced performance trajectory, achieving high fidelity to dialectal forms in the zero-shot setting and enhancing translation quality with the relevant examples. This exceptional capability likely stems from its massive scale, affording better instruction adherence and broader linguistic knowledge~\citep{NEURIPS2020_1457c0d6,JMLR:v25:23-0870}. In the ZS setting with \texttt{DIA-REFINE(M)}, the model achieved a DFS of 0.40 and almost perfect TDR of 0.99. However, its n-gram scores were low with chrF++/BLEU scores of 38.55/22.13, indicating a prioritization of dialectal form over lexical overlap with the reference. The introduction of ICL led to a substantial improvement in translation quality, and with \texttt{DIA-REFINE(M)}, it yielded chrF++/BLEU scores of 61.68/47.22 while maintaining the exceptional TDR of 0.99. This trajectory suggests that in the zero-shot setting, the model initially focused on satisfying the formal requirements of the dialect classifier and subsequently enhanced translation accuracy by leveraging in-context examples.

Our experimental results also revealed that models exhibit varying degrees of responsiveness across the different methods. HyperCLOVAX, Llama-3.1, and EEVE predominantly exhibited a \textbf{False Success} tendency. They demonstrated limited improvement, achieving high n-gram scores but negative DFS scores and TDRs of around 0.40, even when ICL and \texttt{DIA-REFINE(M)} were applied simultaneously.

In contrast, EXAONE-3.5 initially presented a clear case of failure, recording low scores on both n-gram and the proposed metrics. This model, however, exhibited a remarkable improvement following the integration of ICL and \texttt{DIA-REFINE(M)}, with its DFS increasing to 0.27 and TDR to 0.94, along with a corresponding improvement in n-gram scores. These results suggest that the effectiveness of such approaches for dialect translation may rely on the model's inherent linguistic capability.

\subsection{Comparison of DIA-REFINE Variants under Challenging Conditions}\label{sec:6.3}

To evaluate dialect translation under more challenging conditions, we compiled a separate set of difficult samples. Specifically, during the experiments reported in Table~\ref{tab:full_results}, we collected all instances in which every \texttt{DIA-REFINE} method failed on the first attempt, yielding a total of 2,305 samples\footnote{We selected 2,305 samples from an initial pool of 4,500 test instances, generated by evaluating 5 models on 300 samples for each of 3 dialects.}. The results of experiments conducted exclusively on this subset are presented in Table~\ref{tab:difficult_samples}. Under both the ZS and ICL settings, the application of \texttt{DIA-REFINE(S)} and \texttt{(M)} yielded incremental improvements in the proposed DFS and TDR metrics. In particular, TDR increased to 0.09 under the ZS and to 0.28 under ICL setting, demonstrating that relevant examples remain highly influential even for these challenging cases.

It is noteworthy that, even when only the more challenging samples were selected, the proposed metrics exhibited lower scores due to its difficulty, whereas the n-gram-based metrics such as chrF++ and BLEU remained relatively stable. This observation suggests that the high n-gram scores likely stem from models replicating the source sentences rather than accurately generating the intended dialectal translations. Such results underscore the limitations of traditional metrics in faithfully evaluating dialect translation.

In summary, these results demonstrate the superior corrective capability of \texttt{DIA-REFINE(M)}. The framework not only achieves a markedly higher success rate but also exhibits enhanced efficiency. In the ICL setting, it attained the highest TDR with an average of 2.81 attempts, compared to 2.95 attempts for \texttt{DIA-REFINE(S)}. These findings indicate that the multi-candidate strategy constitutes the most effective mechanism for recovering from initial failures and achieving successful dialect translations under challenging conditions.

\section{Conclusion}\label{sec:7}

In this study, we addressed the challenges of dialect translation and evaluation in LLMs. We introduced dialect refinement (\textbf{\texttt{DIA-REFINE}}), a framework that steers models toward producing high fidelity dialect translations by leveraging feedback from our ensemble dialect classifiers. Through an iterative process of translation, verification, and feedback, the proposed framework improves dialect translation success rates and fidelity, with the multi-candidate variant proving to be the most effective. Furthermore, to overcome the limitations of traditional evaluation, we proposed new metrics comprising the DFS and TDR. We demonstrated that these metrics can identify cases of \textbf{False Success}, where n-gram-based scores reward models for defaulting to the standard language, thereby enabling a more faithful evaluation of true dialect translation capability.

Our analysis revealed that the effectiveness of the \texttt{DIA-REFINE} framework depends on the intrinsic capability of the base model. We found that models exhibiting a \textbf{False Success} tendency were less responsive to the proposed framework than those exhibiting \textbf{True Attempt}, and even the model that initially failed showed substantial improvement with the \texttt{DIA-REFINE(M)}. This study not only presents a robust framework for enhancing dialect translation but also establishes a reliable evaluation that facilitates deeper insights into model behavior beyond conventional metrics.

\section{Ethical considerations}

Our study utilizes the `Korean dialect data of middle-aged and elderly speakers' corpus~\citeplanguageresource{aihub:dialectdata}, which may introduce potential bias by not fully representing the linguistic diversity across age groups. Given the use of LLMs, there exist inherent concerns regarding model biases. To mitigate this, we strictly limited their application to dialect translation and did not use them for any other purpose. This paper used datasets from The Open AI Dataset Project (AI-Hub, S. Korea). All data information can be accessed through *AI-Hub (\url{www.aihub.or.kr}).

\section{Limitations}
As this study focuses on a single language and a corpus filtered for longer sentences, its generalizability is constrained; however, we expect that the proposed framework can be readily extended to other languages and corpora. Given that we focused on dialects with pronounced features, additional methods are required to enable models to handle dialects lacking distinctive characteristics. The efficacy of the framework depends on an external classifier that requires labeled data, leaving its performance under data-scarce conditions undetermined. Semantic preservation was not evaluated, which is a common limitation in low-resource settings. Our evaluation relies on automatic metrics, while qualitative aspects such as fluency and naturalness require verification by native speakers, which we plan to address in future work.


\section{Acknowledgements}
This work was supported by the Institute of Information \& Communications Technology Planning \& Evaluation (IITP) grant funded by the Korea government (MSIT) [RS-2021-II211341, Artificial Intelligence Graduate School Program (Chung-Ang University)] and by the National Research Foundation of Korea (NRF) grant funded by the Korea government (MSIT) (RS-2025-00556246).

\section{Bibliographical References}\label{sec:reference}

\bibliographystyle{lrec2026-natbib}
\bibliography{lrec2026-example}

\section{Language Resource References}
\label{lr:ref}
\bibliographystylelanguageresource{lrec2026-natbib}
\bibliographylanguageresource{languageresource}

\clearpage
\appendix
\onecolumn

\section{Detailed Ensemble Performance Results}

\begin{center}
  \begin{minipage}{0.98\textwidth}
    \centering
    \small
    \setlength{\tabcolsep}{6pt}
    \renewcommand{\arraystretch}{1.15}
    \rowcolors{3}{gray!06}{white}

    \begin{tabularx}{\linewidth}{@{}c Y c@{}}
      \hline
      \textbf{Ensemble Size} & \multicolumn{1}{c}{\textbf{Combination}} & \textbf{Accuracy (\%)} \\
      \hline
      2 & KcELECTRA + KoSimCSE-RoBERTa & 94.92 \\
      3 & KcELECTRA + KoELECTRA + KLUE-BERT & 94.88 \\
      3 & KcELECTRA + KoELECTRA + KLUE-RoBERTa & 94.78 \\
      2 & KcELECTRA + KLUE-RoBERTa & 94.72 \\
      2 & KcELECTRA + KLUE-BERT & 94.70 \\
      2 & KcELECTRA + KoELECTRA & 94.68 \\
      3 & KcELECTRA + KoELECTRA + KoSimCSE-RoBERTa & 94.68 \\
      4 & KcELECTRA + KoELECTRA + KoSimCSE-RoBERTa + KLUE-RoBERTa & 94.62 \\
      4 & KcELECTRA + KoELECTRA + KLUE-BERT + KoSimCSE-RoBERTa & 94.56 \\
      4 & KcELECTRA + KoELECTRA + KLUE-BERT + KLUE-RoBERTa & 94.50 \\
      3 & KcELECTRA + KLUE-BERT + KLUE-RoBERTa & 94.46 \\
      5 & KcELECTRA + KoELECTRA + KLUE-BERT + KoSimCSE-RoBERTa + KLUE-RoBERTa & 94.40 \\
      2 & KoELECTRA + KLUE-RoBERTa & 94.38 \\
      2 & KoELECTRA + KoSimCSE-RoBERTa & 94.36 \\
      1 & KcELECTRA & 94.34 \\
      3 & KoELECTRA + KLUE-BERT + KLUE-RoBERTa & 94.22 \\
      3 & KcELECTRA + KoSimCSE-RoBERTa + KLUE-RoBERTa & 94.20 \\
      3 & KoELECTRA + KoSimCSE-RoBERTa + KLUE-RoBERTa & 94.18 \\
      4 & KoELECTRA + KLUE-BERT + KoSimCSE-RoBERTa + KLUE-RoBERTa & 94.16 \\
      3 & KcELECTRA + KLUE-BERT + KoSimCSE-RoBERTa & 94.14 \\
      4 & KcELECTRA + KLUE-BERT + KoSimCSE-RoBERTa + KLUE-RoBERTa & 94.12 \\
      3 & KoELECTRA + KLUE-BERT + KoSimCSE-RoBERTa & 94.04 \\
      1 & KoELECTRA & 93.98 \\
      2 & KoELECTRA + KLUE-BERT & 93.88 \\
      3 & KLUE-BERT + KoSimCSE-RoBERTa + KLUE-RoBERTa & 93.84 \\
      2 & KLUE-BERT + KoSimCSE-RoBERTa & 93.62 \\
      2 & KoSimCSE-RoBERTa + KLUE-RoBERTa & 93.56 \\
      2 & KLUE-BERT + KLUE-RoBERTa & 93.50 \\
      1 & KLUE-RoBERTa & 93.38 \\
      1 & KLUE-BERT & 93.20 \\
      1 & KoSimCSE-RoBERTa & 92.92 \\
      \hline
    \end{tabularx}

    \vspace{4pt}

    \captionsetup{
      type=table,
      width=\linewidth,
      justification=raggedright,
      singlelinecheck=false,
      format=plain,
      margin=0pt,
      skip=6pt
    }
    \captionof{table}{Comparison of all 31 combinations constructed from five Korean pre-trained language models under identical data and hyperparameters, sorted by accuracy. The Ensemble size column indicates the number of base models (1--5).}
    \label{tab:final_ensemble_results}
  \end{minipage}
\end{center}

\clearpage
\section{Prompts}

\begin{center}
\begin{minipage}{\textwidth}
  \begin{promptbox}
You are a translation assistant that translates standard Korean into {DIALECT}.
Output only the translated sentence; do not include any explanations.
  \end{promptbox}

  \vspace{4pt}
  \captionsetup{
    type=figure,
    width=\linewidth,
    justification=centering,
    singlelinecheck=true,
    format=plain,
    margin=0pt,
    skip=6pt
  }
  \captionof{figure}{System instruction used across all settings.}
  \label{fig:prompt-1-system}
\end{minipage}
\end{center}
\vspace{1.5em}   

\begin{center}
\begin{minipage}{\textwidth}
  \begin{promptbox}
Input sentence: {SOURCE}
Translation:
  \end{promptbox}

  \vspace{4pt}
  \captionsetup{
    type=figure,
    width=\linewidth,
    justification=centering,
    singlelinecheck=true,
    format=plain,
    margin=0pt,
    skip=6pt
  }
  \captionof{figure}{Zero-shot prompt template.}
  \label{fig:prompt-2-zero-shot}
\end{minipage}
\end{center}
\vspace{1.5em}   

\begin{center}
\begin{minipage}{\textwidth}
  \begin{promptbox}
Example:
A: {STANDARD_EXAMPLE_1}
B: {DIALECT_EXAMPLE_1}

Example:
A: {STANDARD_EXAMPLE_2}
B: {DIALECT_EXAMPLE_2}

... (total of 10 examples)

Input sentence: {SOURCE}
Translation:
  \end{promptbox}

  \vspace{4pt}
  \captionsetup{
    type=figure,
    width=\linewidth,
    justification=centering,
    singlelinecheck=true,
    format=plain,
    margin=0pt,
    skip=6pt
  }
  \captionof{figure}{In-context learning prompt template.}
  \label{fig:prompt-3-icl}
\end{minipage}
\end{center}
\vspace{1.5em}   

\begin{center}
\begin{minipage}{\textwidth}
  \begin{promptbox}
[Feedback]
- The previous output was classified as {WRONG_LABEL} instead of the target dialect {DIALECT}.
- Please revise the translation to clearly reflect {DIALECT} features.
- Previous output : {PREV_OUTPUT}
  \end{promptbox}

  \vspace{4pt}
  \captionsetup{
    type=figure,
    width=\linewidth,
    justification=raggedright,
    singlelinecheck=false,
    format=plain,
    margin=0pt,
    skip=6pt
  }
  \captionof{figure}{Feedback prompt template. Used in \texttt{DIA-REFINE}; when an output is classified as a non-target dialect, this feedback is prepended to the base prompt (zero-shot or in-context) on the subsequent attempt to request a revision.}
  \label{fig:prompt-4-cil}
\end{minipage}
\end{center}
\vspace{1.5em}   

\begin{center}
\begin{minipage}{\textwidth}
  \begin{promptbox}
The output oscillates between {last_wrong_label} and {wrong_label}.
Please make the {dialect}-specific features more explicit.
  \end{promptbox}

  \vspace{4pt}
  \captionsetup{
    type=figure,
    width=\linewidth,
    justification=raggedright,
    singlelinecheck=false,
    format=plain,
    margin=0pt,
    skip=6pt
  }
  \captionof{figure}{Oscillation prompt template. Used in \texttt{DIA-REFINE}; when the two most recent hypotheses are classified into different non-target dialects, it is inserted into the feedback prompt to flag oscillation and help the model steer toward the target dialect.}
  \label{fig:prompt-cil-oscillation}
\end{minipage}
\end{center}

\end{document}